\documentclass{article}

\usepackage[utf8]{inputenc}
\usepackage[T1]{fontenc}
\usepackage{hyperref}
\usepackage{url}
\usepackage{booktabs}
\usepackage{amsfonts}
\usepackage{amsmath}
\usepackage{amssymb}
\usepackage{nicefrac}
\usepackage{microtype}
\usepackage{xcolor}
\usepackage{graphicx}
\usepackage{subcaption}
\usepackage{algorithm}
\usepackage{algpseudocode}
\usepackage{multirow}
\usepackage{xspace}
\usepackage{float}
\usepackage{authblk}
\usepackage{natbib}
\newcommand{\eggroll}{\textsc{Eggroll}\xspace}
\newcommand{\es}{ES\xspace}
\newcommand{\snn}{SNN\xspace}
\newcommand{\lif}{LIF\xspace}
\newcommand{\R}{\mathbb{R}}

\title{Gradient-Free Training of Spiking Neural Networks via\\
       Low-Rank Evolution Strategies}

\author[1]{Dhruv Patankar}
\author[2]{Sachit Ramesha Gowda}
\affil[1]{\small Shunya Research \\ \texttt{dhruv@shunyaresearch.systems}}
\affil[2]{\small Shunya Research \\ \texttt{sachit@shunyaresearch.systems}}

\begin{document}

\maketitle

\begin{abstract}
Spiking Neural Networks (\snn{}s) offer compelling energy efficiency on
neuromorphic hardware, yet their training remains challenging because the
discrete spike threshold is non-differentiable.
Surrogate-gradient methods sidestep this by approximating the derivative,
but they impose backpropagation infrastructure that is incompatible with
on-chip learning.
Evolution Strategies (\es) are a natural gradient-free alternative, yet
their computational cost scales with the number of parameters, making
them impractical for large weight matrices.

We present a method for training \snn{}s using \eggroll, a low-rank factorisation of \es perturbations
that reduces per-generation memory from $\mathcal{O}(mn)$ to
$\mathcal{O}(r(m{+}n))$.
Combining \eggroll with a Leaky Integrate-and-Fire \snn on N-MNIST,
we demonstrate that gradient-free training achieves \ 79.21\% test accuracy
while reducing per-generation wall-clock time by \ 2.23$\times$ relative
to full-rank \es.
Our results demonstrate EGGROLL is viable for \snn training, with a clear accuracy-speed tradeoff, compatible with training on neuromorphic hardware without surrogate
gradients.
\end{abstract}

\section{Introduction}
\label{sec:intro}

Energy Efficiency is a huge focus of modern AI research. A lot of research is focused on reducing the energy usage of AI architecture. Intel's Loihi ~\citep{davies2018loihi} showed us that SNNs could be a potential solution on real hardware. The dominant paradaigm in current AI is the transformer architecture ~\citep{DBLP:journals/corr/VaswaniSPUJGKP17}. Transformers are trained on GPUs and use hundreds of watts in their training process. While the artificial neural networks with transformers show remarkable performance, they use a significant amount of energy. The human brain, in comparison, is extremely efficient, only using ~20 watts.
\snn{}s attempt to capture this efficiency: neurons integrate inputs
over time and fire only when a membrane potential threshold is crossed,
enabling sparse, event-driven computation on dedicated neuromorphic chips
such as Intel Loihi~\citep{davies2018loihi} and IBM TrueNorth~\citep{merolla2014truenorth}.

\paragraph{The training problem.}
Despite their efficiency during inference, SNNs are still quite difficult to train. The spiking generation function(Heavside step) has a zero gradient almost everywhere. This makes it impossible to do standard backpropagation, since it requires calculating the gradient. The common solution to this is using surrogate gradients ~\citep{neftci2019surrogate}. Surrogate gradients work by replacing the true gradient of a function with an approximation, which can be differentiated. This allows backpropagation to proceed through layers where it would otherwise be impossible. However, surrogate gradients still require autograd infrastructure, which is not compatible with on chip learning on neuromorphic hardware. 

\paragraph{Evolution Strategies.}
\es~\citep{salimans2017es} is a group of loosely bio inspired training methods that are alternatives to backpropagation that do not require gradients to be calculated. In ES, weight vectors are randomly perturbed and then a fitness evaluation is done on these perturbations. However these methods are very computationally expensive, incurring $O(P mn)$ memory for each generation, for a weight matrix of dimensions $m × n$ and population $P$, making it very challenging to scale on larger networks. 

\paragraph{Our contribution.}
We integrate \eggroll~\citep{sarkar2026evolutionstrategieshyperscale} — which replaces each
full-rank perturbation with a low-rank product $\mathbf{AB}^\top$,
$\mathbf{A} \in \R^{m \times r}$, $\mathbf{B} \in \R^{n \times r}$ —
into the \snn training loop.
This reduces per-generation cost to $\mathcal{O}(r(m{+}n))$ while
preserving the gradient-free property.

Concretely, our contributions are:
\begin{enumerate}
  \item we characterize EGGROLL's behavior on non-differentiable spike functions \eggroll with \snn{}s, enabling
        gradient-free training without surrogate approximations
        (Section~\ref{sec:method}).
  \item A rank ablation showing that accuracy remains relatively stable as $r$
        decreases, with a clear efficiency–accuracy Pareto frontier
        (Section~\ref{sec:experiments}).
  \item Comparison against vanilla \es and surrogate-gradient BPTT on N-MNIST. N-MNIST is a native neuromorphic dataset, with no rate-coding approximation used, making this a more principled evaluation than the one done on a static MNIST dataset. (Section~\ref{sec:experiments}).
\end{enumerate}

\section{Background}
\label{sec:background}
\subsection{N-MNIST Dataset}
\label{sec:background_nmnist}
The N-MNIST dataset ~\citep{10.3389/fnins.2015.00437} or Neuromorphic MNIST is a spiking version of the popular MNIST dataset that contains static images of hand drawn numbers. It consists of the same 60,000 training and 10,000 test samples as the original MNIST dataset. It was created by mounting the ATIS sensor on a motorized pan tilt unit and moving it while it recorded MNIST images on an LCD display. The sensor outputs asynchronous events with 2 polarities: ON events (higher brightness) and OFF events (lower brightness). This is a better fit for SNNs than the regular MNIST dataset because the data is already a spike train, so no rate coding approximation is required. Due to this, the SNN processes the actual sensor output instead of the surrogate encoding.

\subsection{Spiking Neural Networks and the LIF Model}
\label{sec:background_snn}

We model each neuron as a Leaky Integrate-and-Fire unit.
At each discrete timestep $t$, the membrane potential $V_t$ evolves as
\begin{equation}
  V_t = \alpha V_{t-1} + \mathbf{w}^\top \mathbf{s}_{t-1},
  \label{eq:lif}
\end{equation}
where $\alpha \in (0,1)$ is the membrane decay constant (leak factor),
$\mathbf{w}$ the synaptic weight vector, and $\mathbf{s}_{t-1}$ the
binary spike vector from the previous layer.
A spike is emitted when $V_t \geq V_\text{th}$, after which $V_t$ resets
to $V_\text{reset} = 0$.
The network output is the spike count (firing rate) over $T$ timesteps.

\subsection{Surrogate Gradient Training}
\label{sec:background_surrogate}

Because the spike function $\Theta(V - V_\text{th})$ is a Heaviside step,
its true derivative is zero almost everywhere.
Surrogate gradient methods~\citep{neftci2019surrogate} substitute a
smooth proxy $\hat{\Theta}'$ — commonly the fast sigmoid — only during
the backward pass, leaving the forward pass unchanged.
This recovers gradient flow at the cost of introducing an approximation
bias and requiring full backpropagation-through-time (BPTT). Implementing this method requires autograd, which as mentioned earlier, is not compatible with on-chip learning, and makes it unsuitable for neuromorphic hardware. 

\subsection{OpenAI Evolution Strategies}
\label{sec:background_es}

\citet{salimans2017es} propose estimating the gradient of expected fitness
$J(\boldsymbol{\theta})$ as
\begin{equation}
  \nabla_{\boldsymbol{\theta}} J \approx
  \frac{1}{2P\sigma}
  \sum_{i=1}^{P} \bigl[F(\boldsymbol{\theta} + \sigma\boldsymbol{\varepsilon}_i)
  - F(\boldsymbol{\theta} - \sigma\boldsymbol{\varepsilon}_i)\bigr]
  \boldsymbol{\varepsilon}_i,
  \label{eq:es_grad}
\end{equation}
where $\boldsymbol{\varepsilon}_i \sim \mathcal{N}(\mathbf{0}, \mathbf{I})$,
$P$ is the population size, and $\sigma$ the perturbation scale.
Antithetic (mirrored) sampling reduces variance with no extra sampling budget.
The estimate is passed to Adam~\citep{kingma2014adam} as if it were a
true gradient. The biggest issue with ES is that its memory cost is $\mathcal{O}(Pmn)$, making it slow for large models.

\subsection{EGGROLL: Low-Rank Evolution Strategies}
\label{sec:background_eggroll}

\citet{sarkar2026evolutionstrategieshyperscale} observe that sampling a full perturbation matrix
$\mathbf{E} \in \R^{m \times n}$ is wasteful: for population $P$, the
perturbation tensor requires $\mathcal{O}(Pmn)$ memory.
\eggroll instead draws
\begin{equation}
  \mathbf{E}_i = \frac{1}{\sqrt{r}} \mathbf{A}_i \mathbf{B}_i^\top,
  \quad
  \mathbf{A}_i \in \R^{m \times r},\;
  \mathbf{B}_i \in \R^{n \times r},
  \label{eq:eggroll}
\end{equation}
normalising by $\sqrt{r}$ so that each entry has unit variance regardless
of rank, without which the variance scales by $r$, rendering the $\sigma$ hyperparameter obsolete. Additionally, EGGROLL reconstructs noise on demand using a counter based deterministic random number generator(RNG), so that the perturbations do not have to be stored in memory. This trick helps EGGROLL work on the billion parameter scale.
The gradient estimate is then reconstructed via the
$(\operatorname{diag}(\mathbf{f})\mathbf{A})^\top \mathbf{B}$ formulation
(§4.2 of \citet{sarkar2026evolutionstrategieshyperscale}), which never materialises individual
perturbation matrices.
Memory per generation drops to $\mathcal{O}(r(m{+}n))$.

\section{Method}
\label{sec:method}

\subsection{Network Architecture}
\label{sec:method_arch}

We use a two-layer \lif network:
\begin{equation}
  2312 \xrightarrow{}
  64  \xrightarrow{\text{LIF}} 10 \xrightarrow{\text{LIF}} \text{output}.
\end{equation}
The input is of size 2312 since N-MNIST has a size of $34 X 34$ pixels, and 2 polarities for each pixel. Each \lif layer shares a common membrane decay $\beta$ . Initial weights are sampled from $\mathcal{N}(0,0.3^2)$.
The output is the mean spike rate over $T$ timesteps, converted to
class probabilities via softmax for fitness evaluation.
Biases are perturbed with independent 1-D Gaussian factors to maintain
equivalence with full-rank \es when $r = \min(m,n)$. Unlike static-image SNNs, the forward pass feeds a different event frame to each timestep not the same image repeated T times. Event counts are clamped to $[0,1]$ to prevent a single pixel dominating membrane potential.

\subsection{EGGROLL Integration}
\label{sec:method_integration}

Algorithm~\ref{alg:eggroll_snn} details the training loop. Instead of backpropogating through the SNN the usual way, the algorithm starts with network weights and creates many perturbed copies of the network which are subsequently run on a minibatch. The performance of each copy is measured and that score is then converted into a gradient like signal which updates the weights with Adam.
The key departure from vanilla \es is the batched forward pass in
Line~\ref{alg:fwd}: all $P$ perturbed networks are evaluated
simultaneously by expanding the data tensor along a population dimension,
avoiding a sequential Python loop over population members.
The factors $\mathbf{A}$, $\mathbf{B}$ are regenerated from a stored
seed rather than cached, keeping GPU memory independent of $P$. Centered rank normalization is done to make the gradient scale unaffected by absolute fitness values across batches. Antithetic sampling is also used to reduce the variance.

\begin{algorithm}[t]
\caption{\eggroll training for an \snn}
\label{alg:eggroll_snn}
\begin{algorithmic}[1]
\Require rank $r$, population $P$, scale $\sigma$, learning rate $\eta$,
         generations $G$, timesteps $T$
\State Initialise weights $\boldsymbol{\theta} = \{\mathbf{W}_1, \mathbf{b}_1,
       \mathbf{W}_2, \mathbf{b}_2\}$ from $\mathcal{N}(0, 0.3^2)$
\For{$g = 1, \ldots, G$}
  \State Sample seed $s$; draw data mini-batch $(\mathbf{X}, \mathbf{y})$
  \State Reconstruct $\mathbf{A}_i, \mathbf{B}_i, \mathbf{c}_i$ from $s$ for $i=1,\ldots,P$
  \State \textbf{Batched forward:} evaluate all $P$ perturbed nets on $\mathbf{X}$ \label{alg:fwd}
  \State Collect antithetic rewards $r_i^+$, $r_i^-$ for each population member
  \State Apply centered rank normalisation to $\{r_i^+, r_i^-\}$
  \State Compute gradient $\hat{\nabla}$ via $(\operatorname{diag}(\mathbf{f})\mathbf{A})^\top \mathbf{B}$
  \State Update $\boldsymbol{\theta} \leftarrow \boldsymbol{\theta} - \eta \cdot \text{Adam}(\hat{\nabla})$
\EndFor
\end{algorithmic}
\end{algorithm}

\subsection{Fitness Function}
\label{sec:method_fitness}

We use negative cross-entropy (log-likelihood) as the fitness signal,
rather than raw classification accuracy:
\begin{equation}
  F(\boldsymbol{\theta}) =
  -\frac{1}{B}\sum_{b=1}^{B}
  \log \frac{\exp(\hat{y}_{b,y_b})}{\sum_c \exp(\hat{y}_{b,c})},
\end{equation}
where $\hat{y}_{b,c}$ is the spike rate for class $c$ on example $b$.
Log-likelihood provides a smooth landscape for \es to navigate; raw
accuracy (a step function over fitness values) gives no gradient signal
between threshold crossings.

\section{Experiments}
\label{sec:experiments}

\subsection{Setup}
\label{sec:exp_setup}

\paragraph{Dataset.}
N-MNIST \citep{10.3389/fnins.2015.00437}: 54,000 training / 6,000 validation / 10,000
test images, generator seed 0. Sensor size $34 X 34$, 2 polarities. No rate coding is used, the event frames are the spiking input.

\paragraph{Baselines.}
\begin{itemize}
  \item \textbf{Vanilla \es}: full-rank Gaussian perturbations,
        sequential evaluation, identical hyperparameters.
  \item \textbf{Surrogate-gradient BPTT}: fast-sigmoid surrogate,
        Adam optimiser, same architecture.
\end{itemize}

\paragraph{Compute.}
All experiments run on a single NVIDIA RTX 3070 Ti.
Wall-clock times are averaged over three seeds.

\subsection{Main Results}
\label{sec:exp_main}

\begin{table}[H]
\centering
\caption{Test accuracy and per-generation wall-clock time.
         Mean $\pm$ std over three seeds.}
\label{tab:main_results}
\begin{tabular}{lccc}
\toprule
Method & Test Acc.\ (\%) & Time / gen (s) & Speedup \\
\midrule
Surrogate BPTT       & 94.03$\pm$0.28 & 31.69 & 0.41$\times$ \\
Vanilla \es          & 76.17$\pm$6.07  & 13.00 & 1.0$\times$ \\
\eggroll $r=1$       & 78.75$\pm$0.58  & 3.41  & 3.81$\times$ \\
\eggroll $r=2$       & 72.27$\pm$12.00 & 5.96  & 2.18$\times$ \\
\eggroll $r=4$       & 79.21$\pm$0.64  & 5.82  & 2.23$\times$ \\
\eggroll $r=8$       & 72.80$\pm$6.08  & 5.88  & 2.21$\times$ \\
\bottomrule
\end{tabular}
\end{table}

The key takeaway from \autoref{tab:main_results} is the significant speedup that EGGROLL provides to ES, for almost no tradeoff in accuracy. Even with a rank of 1, the difference in accuracy is very minute relative to the speedup provided by EGGROLL. Though Surrogate BPTT has a higher accuracy than ES, it is significantly slower. 
\subsection{Rank Ablation}
\label{sec:exp_rank}

\begin{figure}[H]
  \centering
  \begin{subfigure}{0.48\linewidth}
    \includegraphics[width=\linewidth]{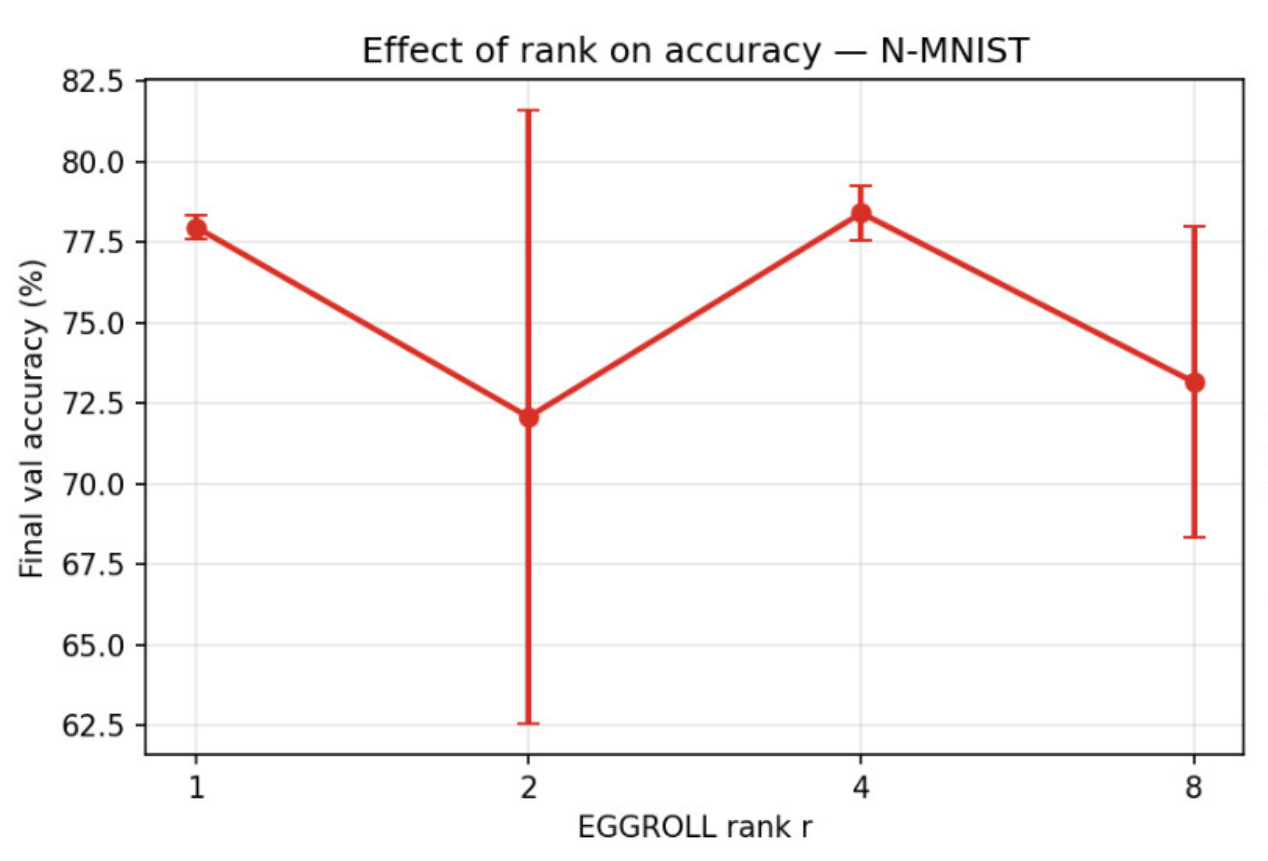}
    \caption{Accuracy vs.\ rank.}
  \end{subfigure}
  \hfill
  \begin{subfigure}{0.48\linewidth}
    \includegraphics[width=\linewidth]{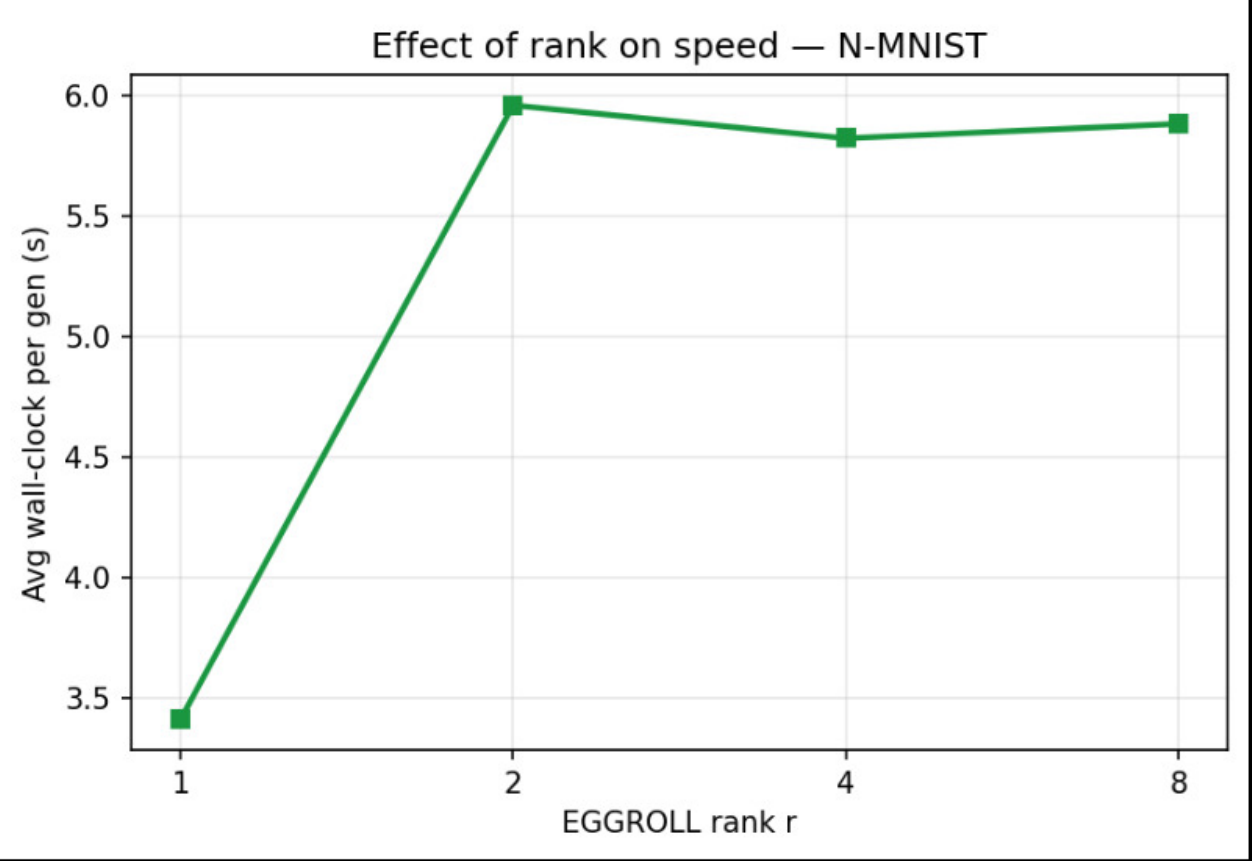}
    \caption{Wall-clock vs.\ rank.}
  \end{subfigure}
  \caption{Rank ablation on N-MNIST.}
  \label{fig:rank_ablation}
\end{figure}

Figure~\ref{fig:rank_ablation} plots validation accuracy and
per-generation wall-clock time as functions of rank $r$.

The accuracy remains relatively stable across ranks, with the best performance occurring at $r=1$ and $r=4$, with larger variance at $r=2$ and $r=8$. Wall clock time is lowest at $r=1$. It increases for $r=2$, and remains relatively stable across the other 2 ranks.

\subsection{Convergence Curves}
\label{sec:exp_convergence}
\begin{figure}[H]
  \centering
  \includegraphics[width=\linewidth]{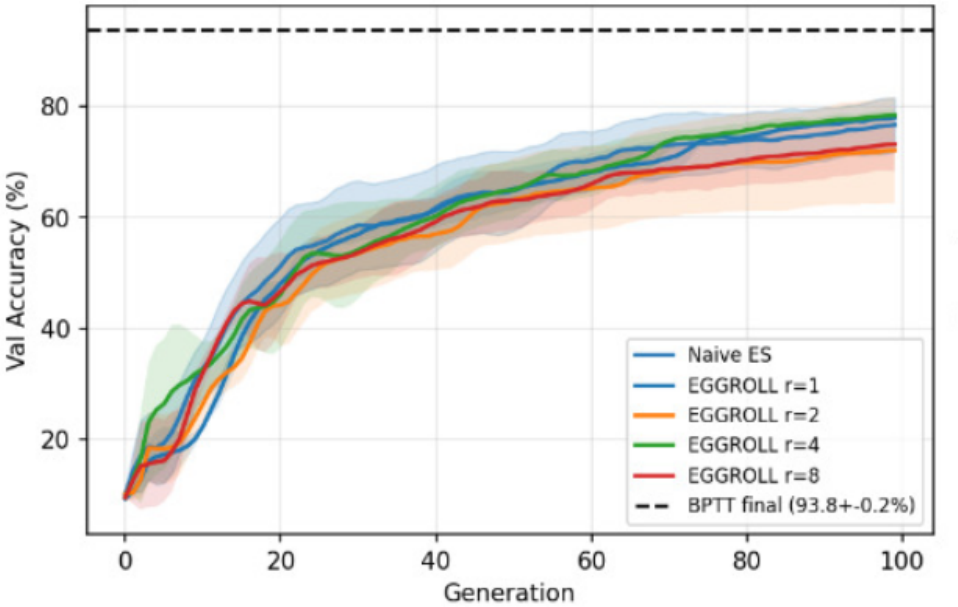}
  \caption{Validation accuracy over generations for all methods (mean
           $\pm$ std, three seeds).
           Dashed line: surrogate BPTT final accuracy.}
  \label{fig:convergence}
\end{figure}

\begin{figure}[H]
  \centering
  \includegraphics[width=\linewidth]{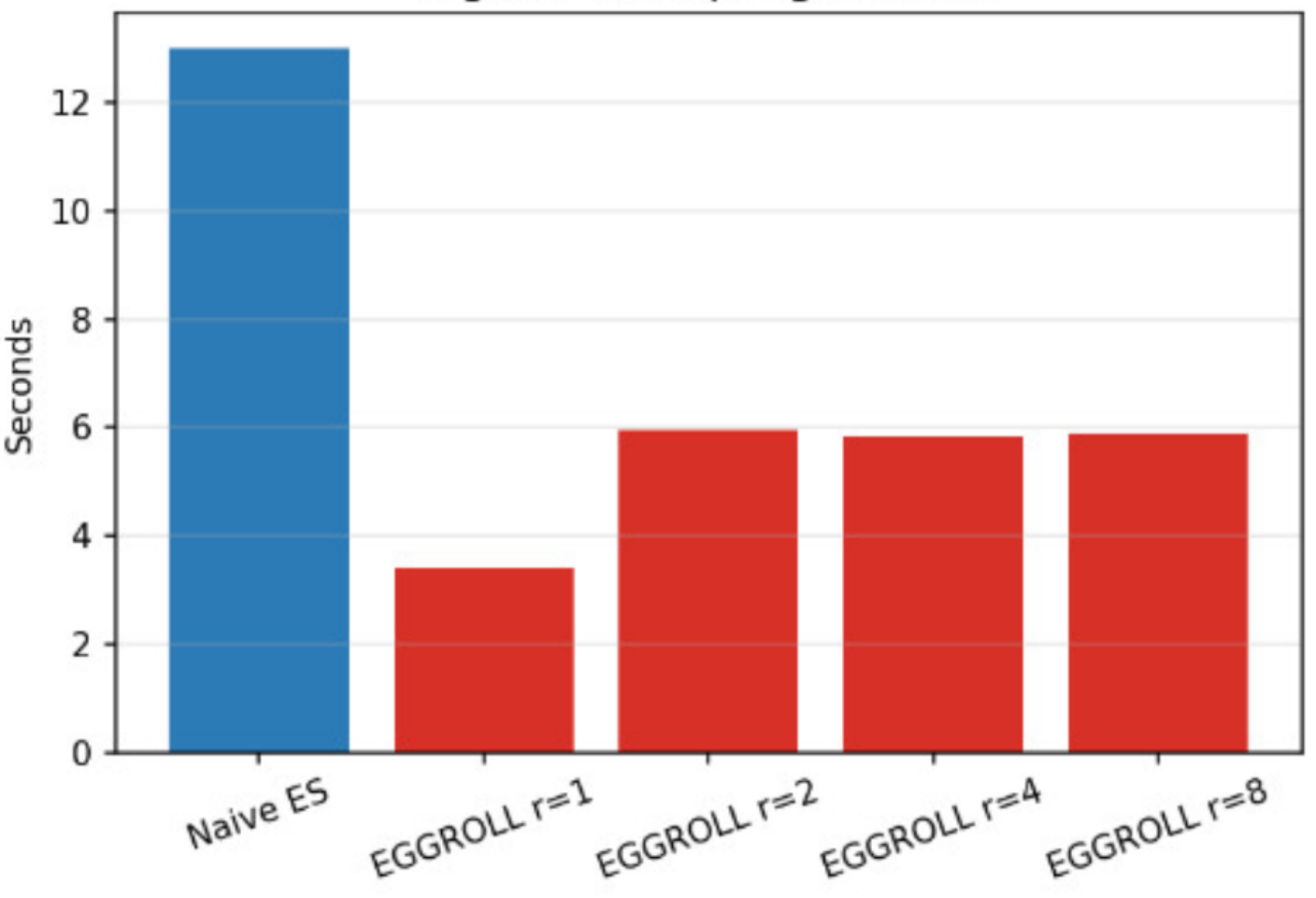}
  \caption{Wall clock time(in seconds) for EGGROLL implementations and vanilla ES}
  \label{fig:time}
\end{figure}

Figure ~\ref{fig:convergence} plots the convergence of different ES implementations(Naive and with EGGROLL) across generations. Figure ~\ref{fig:time} plots the average wall clock time for vanilla ES and EGGROLL.

EGGROLL with $r=4$ converges slightly faster than the other EGGROLL variants, with $r=1$ slightly behind, but overall the difference in convergence speed is modest. The variance bands
(mean $\pm$ std over three seeds) are generally narrow for
$\eggroll$ with small ranks, indicating more stable convergence
than several higher-rank or full-rank baselines. EGGROLL achieves substantially lower wall-clock training time due to its reduced per-generation cost. In particular EGGROLL reaches comparable validation accuracy in fewer seconds than vanilla ES despite similar generation-level convergence behavior.

\section{Discussion}
\label{sec:discussion}

\paragraph{Accuracy gap.}
\eggroll achieves \ 79.21\% accuracy vs.\ \ 94.03\% for surrogate BPTT.
We attribute the gap to our limited compute budget as a result of which we could only run 100 generations for the ES training(naive and EGGROLL). Additionally with ES, the gradient estimate can be noisy because of stochastic perturbations and mini-batch variation. Due to our limited compute budget, we used a lightweight two-layer fully connected SNN, which keeps the parameter count modest, but limits our representational capacity relative to deeper architectures. Per-seed results (Appendix~\ref{app:seeds}) show occasional training collapse at $r=2$, $r=8$, suggesting sensitivity to initialization that warrants further study.

\paragraph{Rank and expressiveness.}
The rank ablation (Figure~\ref{fig:rank_ablation}) shows that even $r=1$
retains substantial accuracy, suggesting the useful gradient directions
lie in a low-dimensional subspace.
This is consistent with the flat minima hypothesis~\citep{hochreiter1997flat}
and with findings on intrinsic dimensionality in neural
networks~\citep{li2018intrinsic}. The ablations are empirical evidence for this in SNNs.

\paragraph{Limitations.}
Our experiments are limited to N-MNIST with a two-layer, fully connected network with no convolutions, and 10 fixed time bins which may discard some temporal resolution. Additionally we have not tested this on real neuromorphic hardware, and have not measured the energy consumption. 
Scaling to CIFAR10-DVS~\citep{10.3389/fnins.2017.00309} and N-Caltech101 ~\citep{10.3389/fnins.2015.00437} is the next step, since these are more complex than N-MNIST. Three seeds is insufficient for definitive rank claims; we report per-seed results in Appendix ~\ref{app:seeds} for transparency.

\section{Related Work}
\label{sec:related}

\paragraph{SNN training methods.}
Spike-Timing-Dependent Plasticity (\textsc{STDP}) is a local Hebbian rule
that requires no global error signal~\citep{bi1998synaptic} but struggles
to match backprop accuracy.
Surrogate gradients~\citep{neftci2019surrogate,bellec2020solution} are
currently the dominant approach for deep \snn{}s, achieving state-of-the-art
accuracy but requiring autograd.
Conversion methods~\citep{cao2015spiking} train an ANN and convert weights,
but incur long simulation times.

\paragraph{Evolution Strategies for neural networks.}
\citet{salimans2017es} showed \es competitive with RL on Atari.
\citet{such2017deep} demonstrated \es on deep networks via compact
perturbation seeds.
PEPG~\citep{SEHNKE2010551} and CMA-ES~\citep{hansen2016cma} are
closely related natural-gradient variants.
\eggroll~\citep{sarkar2026evolutionstrategieshyperscale} is, to our knowledge, the first
explicit low-rank factorisation of \es perturbations shown to scale to
large models; our work is the first to apply it to \snn{}s.

\paragraph{Neuromorphic hardware constraints.}
Loihi~\citep{davies2018loihi} and BrainScaleS~\citep{schemmel2010wafer}
support on-chip learning but only with local rules or very limited
precision.
Gradient-free methods are a natural fit for these constraints, motivating
this line of work.

\paragraph{EGGROLL}
~\citet{sarkar2026evolutionstrategieshyperscale} showed that EGGROLL can work on language models and RNNs. As far as we know, our work is first application of EGGROLL on SNNs, that poses the challenge of non differentiable spikes.

\section{Conclusion}
\label{sec:conclusion}

Training on \snn{}s is challenging due to its discrete spike threshold. We have demonstrated that \eggroll — low-rank Evolution Strategies — can
train \snn{}s on N-MNIST without surrogate gradients or backpropagation,
achieving \ 79.21\% test accuracy at \ 2.23$\times$ the speed of vanilla \es.
The rank $r=4$ offers the
best accuracy-per-second tradeoff in our setup.

\paragraph{Future work.}
The most immediate extension is scaling to more complex neuromorphic benchmarks, like N-Caltech101 ~\citep{10.3389/fnins.2015.00437} and DVS-CIFAR10 ~\citep{10.3389/fnins.2015.00437} with a convolutional SNN.
Beyond accuracy, a key open question is whether \eggroll-trained \snn{}s
produce sparser spike trains than surrogate-gradient methods, which would
directly translate to energy savings on neuromorphic hardware.
Finally, combining \eggroll with local plasticity rules for the early
layers — reserving global \es updates for the readout layer — may offer
the best of both worlds.


\bibliographystyle{plainnat}
\bibliography{references}


\appendix
\section{Variance Analysis Across Seeds}
\label{app:seeds}

\begin{table}[H]
\centering
\caption{Per-seed accuracy and average training time across three seeds.}
\label{tab:seed_variance}
\begin{tabular}{lcc}
\toprule
Method & Test Acc.\ (\%) & Time / gen (s) \\
\midrule
Vanilla ES (seed 1) & 80.66 & 12.41 \\
Vanilla ES (seed 2) & 78.59 & 13.15 \\
Vanilla ES (seed 3) & 69.26 & 13.43 \\
\midrule
$\eggroll$ $r=1$ (seed 1) & 79.27 & 3.04 \\
$\eggroll$ $r=1$ (seed 2) & 78.13 & 3.59 \\
$\eggroll$ $r=1$ (seed 3) & 78.86 & 3.60 \\
\midrule
$\eggroll$ $r=2$ (seed 1) & 79.15 & 6.06 \\
$\eggroll$ $r=2$ (seed 2) & 79.25 & 5.94 \\
$\eggroll$ $r=2$ (seed 3) & 58.42 & 5.88 \\
\midrule
$\eggroll$ $r=4$ (seed 1) & 79.94 & 5.90 \\
$\eggroll$ $r=4$ (seed 2) & 78.98 & 5.82 \\
$\eggroll$ $r=4$ (seed 3) & 78.72 & 5.76 \\
\midrule
$\eggroll$ $r=8$ (seed 1) & 79.78 & 5.75 \\
$\eggroll$ $r=8$ (seed 2) & 69.94 & 5.64 \\
$\eggroll$ $r=8$ (seed 3) & 68.68 & 6.27 \\
\midrule
Surrogate BPTT (seed 1) & 94.35 & 32.09 \\
Surrogate BPTT (seed 2) & 93.88 & 31.67 \\
Surrogate BPTT (seed 3) & 93.86 & 31.32 \\
\bottomrule
\end{tabular}
\end{table}

\section{EGGROLL Gradient Derivation}
\label{app:derivation}

Following \citet{sarkar2026evolutionstrategieshyperscale} §4.2, the gradient estimate for
weight matrix $\mathbf{W}$ under low-rank perturbations is:
\begin{align}
  \hat{\nabla}_{\mathbf{W}} J
  &= \frac{1}{2P\sigma\sqrt{r}}
     \sum_{i=1}^{P} f_i \mathbf{A}_i \mathbf{B}_i^\top \notag \\
  &= \frac{1}{2P\sigma\sqrt{r}}
     \bigl(\operatorname{diag}(\mathbf{f})\,\mathbf{A}\bigr)^\top \mathbf{B},
  \label{eq:eggroll_grad}
\end{align}
where $\mathbf{A} \in \R^{P \times m \times r}$ and $\mathbf{B} \in \R^{P \times n \times r}$
are stacked factor tensors, and $f_i = r_i^+ - r_i^-$ is the antithetic
rank difference.
Equation~\eqref{eq:eggroll_grad} requires only two batched matrix
multiplications, never materialising the $P \times m \times n$ perturbation tensor.

\end{document}